\documentclass[a4paper, 10pt, conference]{ieeeconf}      

\usepackage{amsmath,amsfonts}
\usepackage[noend]{algpseudocode}
\usepackage{algorithmicx}
\usepackage{algorithm}
\usepackage{url}
\usepackage{graphicx}
\usepackage{cite}
\usepackage[font=small]{caption}
\usepackage{subcaption}

\IEEEoverridecommandlockouts                              

\title{\LARGE \bf
Robust Route Planning with Distributional Reinforcement Learning in a Stochastic Road Network Environment}

\author{Xi Lin$^{1}$, Paul Szenher$^{1}$, John D. Martin$^{2}$, Brendan Englot$^{1}$
\thanks{$^{1}$Xi Lin, Paul Szenher and Brendan Englot are with the Department of Mechanical Engineering, Stevens Institute of Technology, 1 Castle Point Terrace, Hoboken, NJ 07030, USA
{\tt\small \{xlin26,pszenher,benglot\}@stevens.edu}}%
\thanks{$^{2}$John D. Martin is with the Department of Computing Science, University of Alberta, 116 St. and 85 Ave., Edmonton, Alberta T6G 2R377, Canada
{\tt\small jmartin8@ualberta.ca}}%
}

\begin{document}

\maketitle
\thispagestyle{empty}
\pagestyle{empty}

\begin{abstract}
Route planning is essential to mobile robot navigation problems.
In recent years, deep reinforcement learning (DRL) has been applied to learning optimal planning policies in stochastic environments without prior knowledge.
However, existing works focus
on learning policies that maximize the expected return,  the performance of which can vary greatly when the level of stochasticity in the environment is high.
In this work, we propose a distributional reinforcement learning based framework that learns return distributions which explicitly reflect environmental stochasticity. 
Policies based on the second-order stochastic dominance (SSD) relation can be used to make adjustable route decisions according to user preference on performance robustness.
Our proposed method is evaluated in a simulated road network environment, and experimental results show that our method is able to plan the shortest routes that minimize stochasticity in travel time when robustness is preferred, while other state-of-the-art DRL methods are agnostic to environmental stochasticity.
\end{abstract}


\section{Introduction}

Endowing mobile robots with the ability to autonomously navigate has been viewed as a major challenge to realizing their full potential as autonomous embodied systems \cite{roy2021machine, yurtsever2020survey}.
In an autonomous navigation task, a mobile robot is typically asked to navigate collision-free from a start point to a goal point, sometimes without an accurate or complete prior map, and it is challenging to efficiently plan routes without prior knowledge.

Reinforcement Learning (RL) offers a way for embodied systems to acquire high-performance policies to operate in stochastic environments through trial-and-error learning \cite{sutton2018reinforcement}. 
Given no prior information describing the environment, a RL agent can learn 
by directly interacting with the environment.
Deep Reinforcement Learning (DRL) combines RL with deep neural network architectures, and has been widely used to solve practical problems with high-dimensional sensory inputs in recent years.
One of the earliest examples, DQN \cite{mnih2015human}, was able to play Atari 2600 games to superhuman levels.
DRL methods have also shown the ability to learn desired route planning policies in unknown stochastic environments, which can be efficiently executed via neural network inference at run-time. 

\begin{figure}[t]
    \centering
    \includegraphics[width=0.95\linewidth]{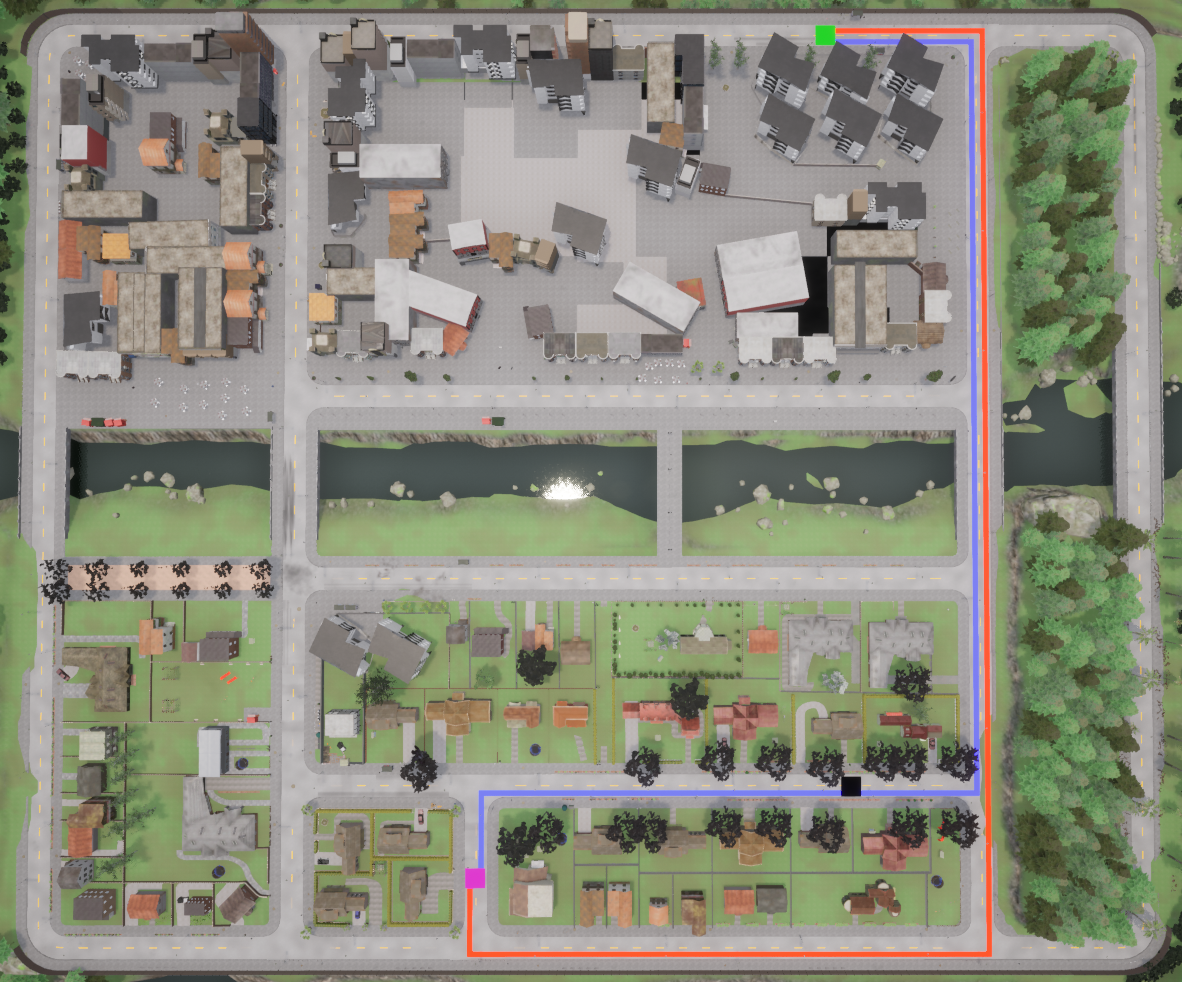}
    \caption{\textbf{Example routes in a simulated urban environment}. Green, magenta and black blocks represent the start, goal, and a crosswalk with stochastic delay. The blue line is the shortest length route.}
    \label{fig:intro}
    \vspace{-6mm}
\end{figure}

Existing DRL based route planners select the route that maximizes the expectation of accumulated rewards (e.g., the blue path in Figure \ref{fig:intro}), which may be associated with high variance and suffer from unstable performance across trials when the operating environment is highly stochastic.
A possible solution is Distributional Reinforcement Learning (Distributional RL) \cite{bellemare2017distributional}, which aims at learning return distributions and enables various policies beyond expectation maximization.
Recent works in this area (described below) have used Distributional RL to learn mobile robot controllers that can achieve safe navigation performance by avoiding actions with highly uncertain outcomes.

In this paper, we propose a route planning framework that learns environmental stochastitcity with distributional RL, and is capable of selecting a robust route with execution policies based on the second-order stochastic dominance (SSD) relation.
The performance of our framework is demonstrated in simulated urban road network environments where travel time stochasticity exists in certain states.
While state-of-the-art RL methods are agnostic to travel time stochasticity, our framework could plan routes that circumvent those states when robustness is preferred (e.g., the red path in Figure \ref{fig:intro}). 
The code of our approach is available at \url{https://github.com/RobustFieldAutonomyLab/Stochastic_Road_Network}.

The rest of this paper is organized as follows: 
Section \ref{sec:rel-work} introduces related works about RL based route planners and distributional RL based mobile robot controllers; 
Section \ref{sec: problem setting} introduces the RL problem setup; Section \ref{sec: robust planning} introduces our proposed robust learning and execution framework based on distributional RL; 
Section \ref{sec: environment} introduces the road network environments we build for studying the robust route planning problem; 
Section \ref{sec: experiment results} shows route planning experiment results and analysis;
Section \ref{sec:conclusion} concludes the paper and comments on plans for future work.


\section{Related Work}
\label{sec:rel-work}


Deep reinforcement learning (DRL) has been used to develop solutions to route planning problems in unknown stochastic environments.
Qian et al. \cite{qian2019deep} considered the charging station navigation problem for electric vehicles in an urban environment with unknown uncertainty in traffic conditions, price and waiting time at different charging stations, and proposed a DRL based planner that could select a goal station and route that minimize travel time and charging cost.  
Zhang et al. \cite{zhang2020route} also worked on the route planning problem for electric vehicles, but with an emphasis on the power management issue; 
they used reinforcement learning to train a route planner that could efficiently plan paths that minimize overall energy consumption.
Geng et al. \cite{geng2021deep} proposed a DRL based route planner 
that minimizes travel time by predicting pedestrian flows in congested environments;
the proposed planner outperforms A* when congestion occurs.
Koh et al. \cite{koh2020real} proposed a DRL based real-time navigation system operating in complex city maps with varying traffic conditions;
the system re-plans the route for the ego vehicle given its surrounding instant traffic information, achieving better performance in travel time than dynamic-Dijkstra and dynamic-A*.   
Basso et al. \cite{basso2022dynamic} studied dynamic route planning for electric delivery service vehicles where customer requests and energy consumption are stochastic, proposing a DRL based solution that learns to anticipate future customer requests and energy consumption, planning routes for delivery and battery charging that minimize energy consumption and prevent battery depletion.

Distributional reinforcement learning (Distributional RL) has recently been used to learn risk-aware control policies in mobile robot navigation tasks. Choi et al. \cite{choi2021risk} focused on the indoor navigation of a mobile ground robot, and trained a distributional RL agent that could learn the collision risk of control commands (linear and angular velocity); the proposed method outperforms  traditional RL in the collision avoidance success rate.
Kamran et al. \cite{kamran2021minimizing} designed a distributional RL based autonomous vehicle controller for automated merging and crossing at occluded intersections, where other vehicles' behaviors are uncertain; the proposed method exhibits adjustable performance that balances travel time and driving comfort according to user preference at run-time.  
Liu et al. \cite{liu2022adaptive} proposed an adaptive risk-aware control policy for a micro drone navigation task; the algorithm automatically adjusts the threshold of Conditional Value at Risk (CVaR) according to the level of environmental uncertainty during the learning of a distributional RL agent, which shows robust navigation performance without human effort in parameter tuning.   
Given that distributional RL has been proven to be effective in learning risk-aware policies, we use it to develop a high-level route planning framework that is robust to uncertainties in the environment. 

\section{Problem Setting}
\label{sec: problem setting}

Using the standard problem setting for RL, the interaction between the agent and the environment is formalized as a Markov Decision Process $(\mathcal{S},\mathcal{A},\mathcal{P},R,\gamma)$, where $\mathcal{S}$, $\mathcal{A}$, and are the set of states and actions. 
At each time step $t$, the agent receives an observation of the current state $s_t \in \mathcal{S}$, and decides an action $a_t \in \mathcal{A}$. 
It causes the environment transition to the state $s_{t+1}\sim \mathcal{P}(\cdot|s_t,a_t)$ , and the agent receives an observation of $s_{t+1}$, as well as a reward $r_{t+1} = R(s_{t+1},a_{t+1})$.
A common way to find an optimal policy in RL is to maximize the expected future return $Q^\pi$.
\begin{equation}
  Q^\pi(s, a) = \mathbb{E}_\pi[\sum_{k=0}^\infty \gamma^k r_{t+k+1}| s_t =s, a_t=a].
\end{equation}

It's also known as the action-value function, which starts with a state-action pair $(s_t,a_t)$ and follows the policy $\pi$ thereafter.
Discount factor $\gamma\in [0,1)$ controls the effect of future rewards. 
\begin{equation}
    \label{eqn:Bellman}
    Q^\pi(s,a) = \mathbb{E}[R(s,a)] + \gamma\mathbb{E}[Q^\pi(s',a')]
\end{equation}
\begin{equation}
    \label{eqn:Bellman optimality operatoer}
    \mathcal{T}Q(s,a) := \mathbb{E}[R(s,a)] + \gamma\mathbb{E}[\max_{a'}Q(s',a')]
\end{equation}

The relation between $Q^\pi$ and its successors satisfies the Bellman equation \eqref{eqn:Bellman}, where $s'\sim \mathcal{P}(\cdot|s,a)$, and $a'\sim \pi(\cdot|s')$. An optimal policy could be solved using the Bellman optimality operator \eqref{eqn:Bellman optimality operatoer}. 
\begin{equation}
    \label{eqn:Distributional bellman}
    Z^\pi(s,a) \overset{D}{=} R(s,a) + \gamma Z^\pi(s',a') 
\end{equation}
\begin{equation}
    \label{eqn:Distributional bellman optimality operator}
    \mathcal{T}Z(s,a) :\overset{D}{=} R(s,a) + \gamma Z^\pi(s',\text{argmax}_{a'} \mathbb{E}[Z(s',a')]) 
\end{equation}

Instead of the expected return, Distributional RL algorithms  \cite{bellemare2017distributional} focus on the return distribution, which satisfies the distributional Bellman equation \eqref{eqn:Distributional bellman}, where $Z^\pi(s,a)$ is a random variable that satisfies $Q^\pi(s,a) = \mathbb{E}[Z^\pi(s,a)]$. Similarly, the distributional Bellman optimality equation \eqref{eqn:Distributional bellman optimality operator} is used in this case.
This permits the estimation of a collection of outcomes that could result from the stochasticity inherent in the agent's domain.
For instance, if the reward encodes a robot's transition time, then its return distribution reflects the spread in total travel times that result from taking different routes whose random traffic conditions vary.

\section{Robust Learning and Execution Framework}
\label{sec: robust planning}

\begin{figure}[t]
    \centering
    \includegraphics[width=0.9\linewidth]{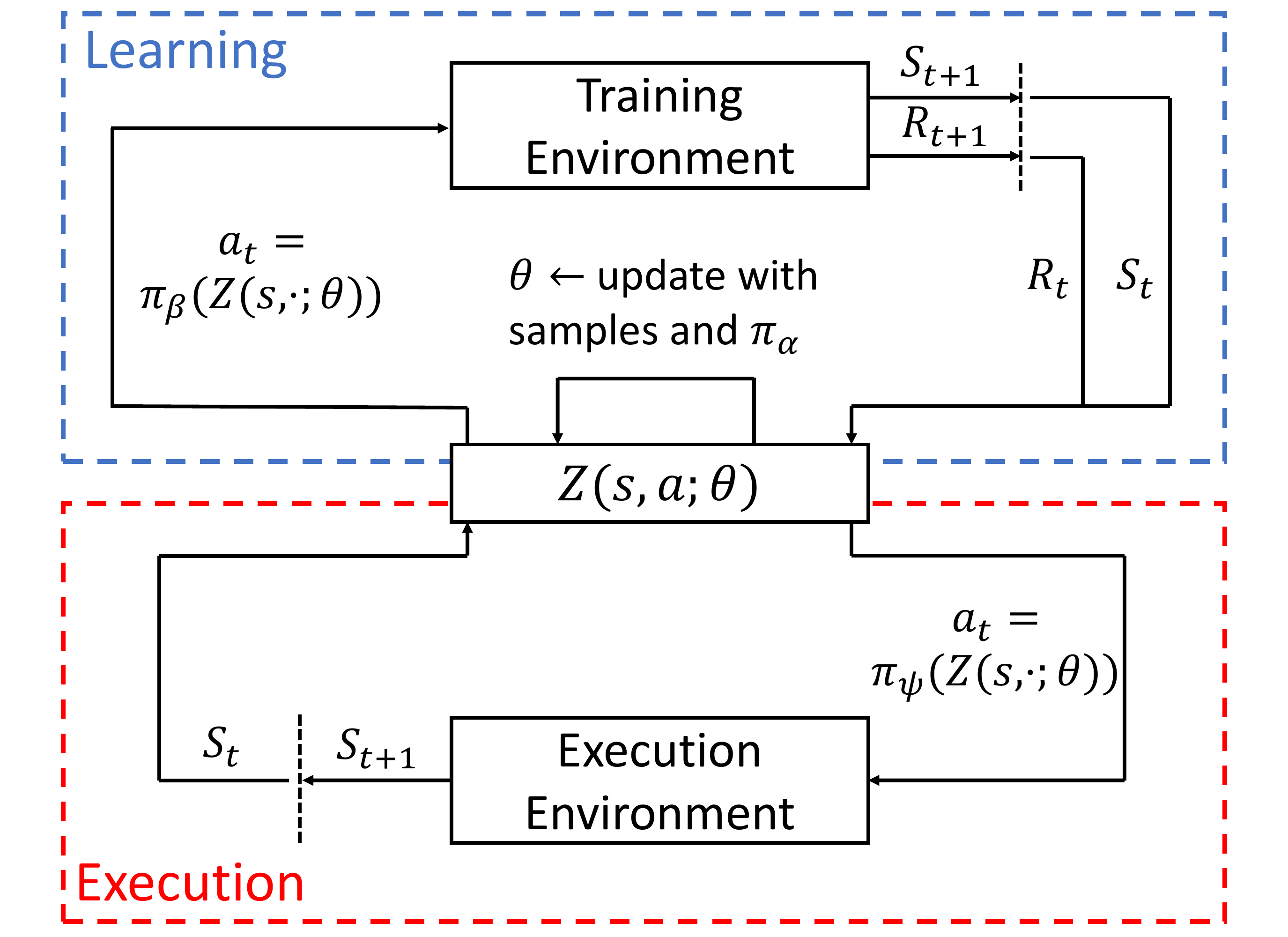}
    \caption{\textbf{Robust Learning and Execution framework}. 
    }
    \label{fig:summary}
    \vspace{-4mm}
\end{figure}

Our proposed robust learning and execution framework is illustrated in Figure  \ref{fig:summary}.
In the learning process, the agent interacts with the environment according to the behavior policy $\pi_\beta$, and updates the return distribution $Z(s,a;\theta)$ with the learning policy $\pi_\alpha$. During the execution process, the agent decides the action to take with the learned return distributions $Z(s,a;\theta)$ and the execution policy $\pi_\psi$.

Consider an example when the learning and execution policies are the same.
In this case, the robot would learn for some period of time, stop using its behavior policy, then start using its execution policy.
In settings where robustness is desired, however, it can be beneficial to use a distinct execution policy to impose additional robustness preferences on decisions.

\subsection{Learning}
We use QR-DQN 
\cite{dabney2018distributional} to learn return distribution $Z(s,a)$, which is approximated with a uniform distribution of $N$ Diracs, $Z_\theta(s,a)$, where $\theta_i$ is the value corresponding to the quantile $\hat{\tau}_i = (\tau_{i-1}+\tau_i)/2$ for $1\leq i\leq N$.
\begin{equation}
    Z_\theta(s,a)=\frac{1}{N}\sum_{i=1}^N \delta_{\theta_i(s,a)}
\end{equation}
\begin{equation}
    \begin{array}{c}
        \rho_\tau^\kappa(u) = |\tau-\delta_{\{u<0\}}|(\mathcal{L}_\kappa(u)/\kappa), \\[5pt]

        \text{where }\mathcal{L}_\kappa(u) = \left\{
            \begin{array}{lr}
                 \frac{1}{2}u^2, &\text{if}\ |u|\leq\kappa \\
                 \kappa(|u|-\frac{1}{2}\kappa), & \text{otherwise} 
            \end{array}
        \right.
    \end{array}
\end{equation}
During each training step, $\{\theta_i(s,a)\}$ are optimized by performing stochastic gradient decent on the quantile regression loss $L_{QR}$, which is computed with quantile TD error $\delta_{ij}$ and quantile Huber loss $\rho_\tau^\kappa$.
\begin{equation}
\label{eqn:quantile_TD}
    \delta_{ij} = r + \gamma \theta_j(s',\pi_{\alpha}(s'))-\theta_i(s,a)
\end{equation}
\begin{equation}
    L_{QR} = \sum_{i=1}^N \mathbb{E}_j[\rho_{\hat{\tau}_i}^\kappa(\delta_{ij})]
\end{equation}
We use the greedy policy as the learning policy in optimization.
\begin{equation}
\label{eqn:tlp}
    \pi_{\alpha}(s)=\text{argmax}_a \mathbb{E}_{z\sim Z_\theta(s,a)}[z]
\end{equation}

\subsection{Execution}
In stochastic environments where a robot can experience multiple and potentially negative outcomes, it can be beneficial to impose an additional preference over outcomes so they are robust. 
Robustness here is specified with the dispersion of a given return distribution, and in the past it has been measured with statistics of the return distribution such as the variance \cite{tamar2013temporal} or Conditional Value at Risk (CVaR) \cite{keramati2020being}.
Martin et al. \cite{martin2020stochastically} showed that policies based on the second-order stochastic dominance (SSD) relation are aggregations of CVaR policies, and these can lead to optimal behaviors that favor robustness when multiple solutions exist.

The SSD relation is defined using distribution functions and compared over the set of their realizable values.
Consider two expected returns, $G$ and $G'$, respectively induced by actions $a$ and $a'$.
We say that $G$ stochastically dominates $G'$ in the second order when their integrated CDFs, $F^{(2)}(z):=\int_{-\infty}^z F(x)dx$, satisfy the following equation, and we denote the relation $G\succeq_{(2)} G'$:
\begin{align}\label{eq:ssd}
        G\succeq_{(2)} G'\Longleftrightarrow F^{(2)}_G(z) \leq F^{(2)}_{G'}(z), \ \forall \ z \in \mathbb{R}.
\end{align}
The function $F^{(2)}$ defines the frontier of what is known as the \textit{dispersion space}, whose volume reflects the degree to which a random variable differs from its expected value.
Outcomes that are disperse have more uncertainty and are less robust when it comes to realizing consistent outcomes.
Indeed, a fundamental result from expected utility theory states that rational risk-averse agents prefer outcome $G$ to $G'$ when $G\succeq_{(2)} G'$ \cite{dentcheva2006inverse}.

We consider the SSD policy as an execution policy, which is based on the SSD relation. 
In settings where there is not an exact tie among the 
actions offering the largest expected returns, the SSD policy will be equivalent to the standard greedy policy.
Otherwise, the SSD policy selects the action that is second-order stochastically dominant.
Here we also introduce a relaxation to the SSD policy that permits uncertainty to be considered even when there is not an exact tie.
The Thresholded SSD policy looks at the action gap between the top two optimal actions.
When this is less than a certain threshold they will be considered equivalent, and the tie will be broken with a comparison of their second central moments, which eliminates the difference in expected values and is consistent with the exact SSD policy---namely, the action that induces the smaller second central moment will be preferred.
Details of the SSD and Thresholded SSD policies are shown in Algorithm \ref{Alg:SSD policy}.

\begin{algorithm}[th!]
\begin{algorithmic}
\Require use\_thres , ssd\_thres
\For{all $a\in \mathcal{A}$}
    \State $Q_a = \frac{1}{N}\sum_{i=1}^N\theta_i(s,a)$
\EndFor
\State // \textit{Identify top-2 actions with largest Q values}
\State $a_1 = \text{argmax}_a \{Q_a\}$,\ $a_2 = \text{argmax}_a \{Q_a \backslash Q_{a_1}\}$
\If{use\_thres is \textbf{true}}
    \State // \textit{Thresholded SSD}
    \If{$Q_{a_1}-Q_{a_2}>$ ssd\_thres}
        \State \textbf{return} $a_1$
    \Else
        \State $V_a = \frac{1}{N}\sum_{i=1}^N (\theta_i(s,a)-Q_a)^2,\ a\in\{a_1,a_2\}$
        \State \textbf{return} $\text{argmin}_a V_a$
    \EndIf
\Else
    \State // \textit{Exact SSD}
    \If{$Q_{a_1}>Q_{a_2}$}
        \State \textbf{return} $a_1$
    \Else
        \State $V_a = \frac{1}{N}\sum_{i=1}^N \theta_i(s,a)^2,\ a\in\{a_1,a_2\}$
        \State \textbf{return} $\text{argmin}_a V_a$
    \EndIf
\EndIf

\end{algorithmic}
\caption{SSD and Thresholded SSD policy}
\label{Alg:SSD policy}
\end{algorithm}

\begin{figure*}[t]
    \centering
    \includegraphics[width=0.8\linewidth]{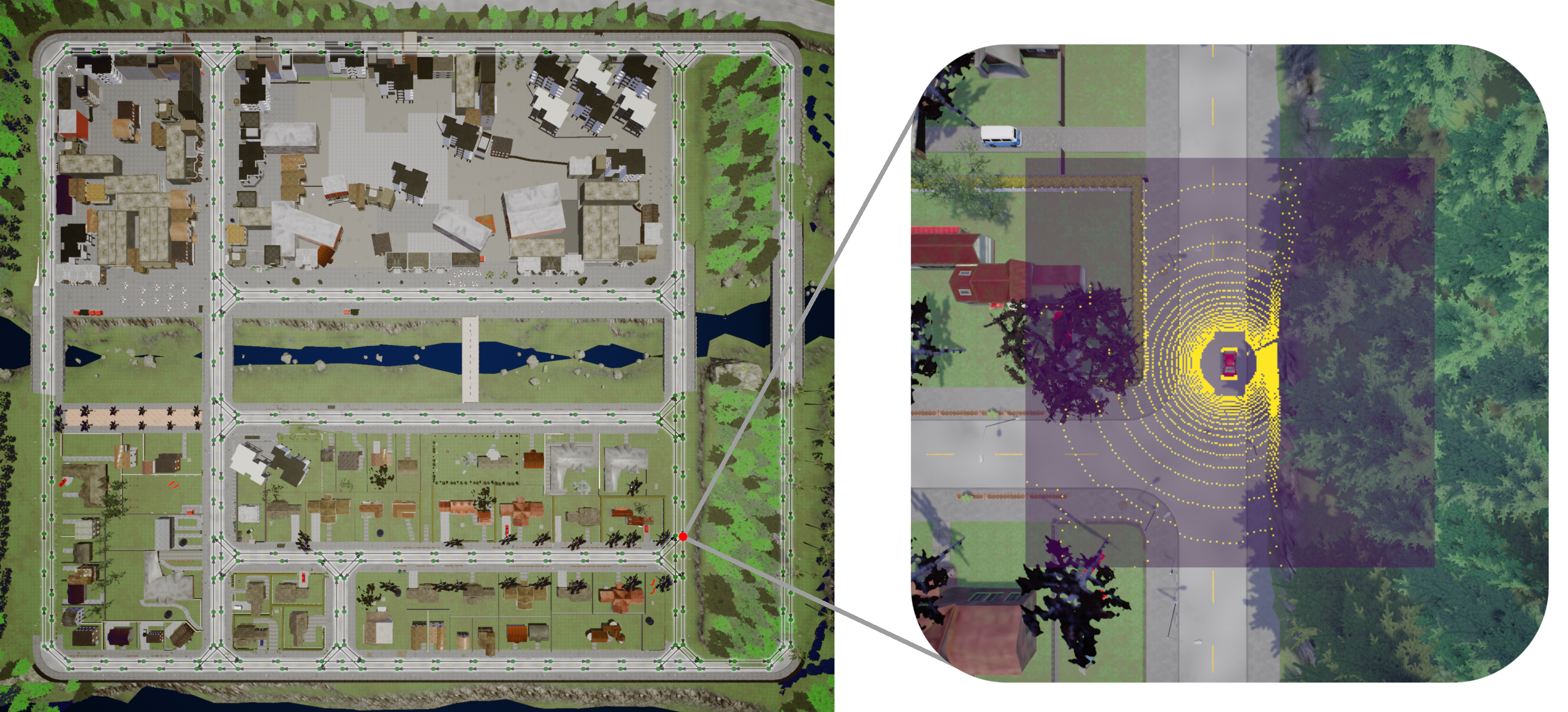}
    \caption{\textbf{Left: CARLA Town01 map and the extracted graph}. Discrete states and actions are shown as green dots and black arrow-head lines. \textbf{Right: an observation sample}. A learning system observes an occupancy image whose pixels correspond to the presence of a LIDAR return within a specific block of space. The image is overlaid on top of the
      corresponding 3D scene.}
    \label{fig:observation}
    \vspace{-4mm}
\end{figure*}

\section{The Stochastic Road Network Environment}
\label{sec: environment}

To examine the performance of the proposed framework on the urban driving route planning problem, we develop a simulation environment called the Stochastic Road Network Environment.  
The environment is built upon map topology and simulated sensory data originating from the CARLA autonomous vehicle simulator \cite{dosovitskiy2017carla}.
We select two maps from CARLA (\texttt{0.9.6} version), \texttt{Town01} and \texttt{Town02}. 

Each map in the proposed environment is procedurally generated.
Starting from the continuous OpenDRIVE maps provided alongside CARLA, maps are extracted through a discretization process that produces a graph approximating the route-level topology of the original CARLA maps.
The waypoint topology is initially represented in the OpenDRIVE road description format \cite{dupuis2010opendrive}---an industry-standard description language of road networks for use in autonomous vehicle simulation.  
The generation procedure produces maps with an offline data pre-processing script, which communicates with the CARLA simulator and generates static map graphs and observational data.
For instance, the graph extracted for the CARLA Town01 road network is visualized in Figure \ref{fig:observation}. 
Given the extensibility of the CARLA simulator framework, the proposed
environment generation method can be extended to cover any custom-made
CARLA maps that include an underlying OpenDRIVE map graph.  

\subsection{State and Action Space}

Each map in the proposed environment is a discrete and directed road network graph, where the state space is the set of all vertices in the graph representing all possible locations that the robot could visit.

The available actions at each state are determined by a discrete map of candidate state
transitions, which correspond to edges in the graph.
The dimension of the action space in the map is the largest number of available actions in any state.
At states where the graph topology does not allow for
use of the full action space (i.e., in the case of a straightaway,
where there is only one valid direction of travel) the remainder
of the action space corresponds to the loopback action, whereby the
agent re-transitions into its current state. State transitions are
fully deterministic, meaning every state-action pair yields a
transition to a deterministic subsequent state.


\subsection{Observation}
The CARLA simulator provides access to an array of rich sensor
data-types.  In this work, rich 3D LIDAR data is extracted from the
CARLA simulator and post-processed into a form more amenable to
usage with reinforcement learning methods.

First, raw returns from a 32-beam, \(360^\circ\) FoV simulated 3D
LIDAR are extracted at each state location in the original CARLA
environment. Next, this data is collected into a simple occupancy-map
format on a per-state-basis, whereby the number of LIDAR returns per
grid-cell are counted and stored in a 3D grid structure.  
This grid structure 
posesses a dimensionality of \(255 \times 255 \times 3\), and a size
of \(50\ m \times 50\ m \times 8.4\ m\), with individual voxels of
size \(0.2m\ \times 0.2m\ \times 2.7m\).  Finally, a single
2D slice of this grid structure (\(255 \times 255\)) is extracted for
use as the environment observation at each environment state.  The
final observation is binarized, yielding a 2-dimensional matrix of
size \(255 \times 255\).  Each cell is populated with a value of \(1\)
if a non-zero number of LIDAR returns were present within the
corresponding occupancy voxel, and \(0\) otherwise. A representation
of the resulting observation format is shown in Figure
\ref{fig:observation}.

When the learning agent reaches a state, it perceives the corresponding observation matrix at that location.  
In this paper, we
are focused on studying how route planning policies are affected by the
environmental stochasticity induced by crosswalks where a vehicle may
need to stop for pedestrians. 
Thus, observation noise is not modeled,
and observation data is non-stochastic, with each binary matrix
representing a static, unchanging observation returned from a
specified state. Subsequent revisiting of a given state will always
yield the same observation.

\subsection{Reward}

Rewards are designed to reflect the travel time cost of passing states, and they are summarized in Equation \eqref{eqn:rewards}.
As most vertices are generated with fixed distances to neighbouring vertices, a fixed base action reward is present at all states to reflect the normal travel time cost, yielding a deterministic reward of \(-r_{base}\).  
In addition to this fixed action
reward, there exist three categories of reward modifiers: goal states, crosswalk states, and loopback actions.

Goal states, denoted as $\mathcal{S}_G$, are destinations of the agent, and reaching a goal state results in the end of a route and an additional reward $+r_{base}$. Hence, the total reward of reaching a goal state is 0.  

Crosswalk states, denoted as $\mathcal{S}_C$, are the sole source of stochasticity within the environment, where travel time varies.
We impose an additional reward sampled from a zero-mean truncated Gaussian distribution with a standard deviation of \(1\) and threshold
bounds of \(-r_{base}\) to \(+r_{base}\). Thus the total reward of reaching a crosswalk state range from \(-2r_{base}\) to \(0\).

Loopback actions are actions which result in the agent
remaining in its current state.  Loopback actions may incur an additional penalty \(r_{loopback}\), resulting in an effective loopback action reward of \(-(r_{base} + r_{loopback})\).
\begin{equation}
\label{eqn:rewards}
    r_t = \left\{
        \begin{array}{cc}
            0, & \text{if } s_t \in \mathcal{S}_G \\
            N(-r_{base},1)\ \&\ r_t\in [-2r_{base},0], & \text{if } s_t \in \mathcal{S}_C \\
            -(r_{base}+r_{loopback}), & \text{if } s_t = s_{t-1} \\
            -r_{base}, & \text{otherwise}
        \end{array}
    \right.
\end{equation}

\section{Experimental Evaluation}
\label{sec: experiment results}
Using the Stochastic Road Network Environment, we study the route planning performance of our proposed framework, as well as that of state-of-the-art RL algorithms for comparison. 
Specifically, in section \ref{sec:baselines}, we investigate the effect of different learning rates on the performance of selected state-of-the-art methods, and report our hyperparameter selection. In section \ref{sec:Robust RL}, we apply the proposed framework to the route planning task, and compare its performance with others.

\subsection{Baselines Study}
\label{sec:baselines}
\begin{figure}
    \centering
        \begin{subfigure}[b]{0.49\columnwidth}
                \centering
                \includegraphics[width=\linewidth]{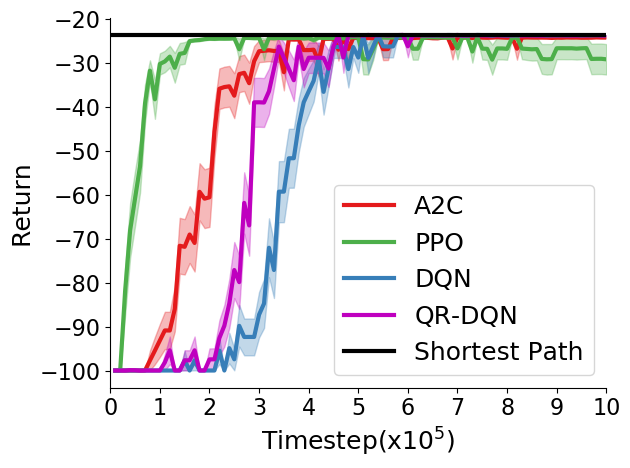}
                \caption{Town01}
        \end{subfigure}
        \begin{subfigure}[b]{0.49\columnwidth}
                \centering
                \includegraphics[width=0.96\linewidth]{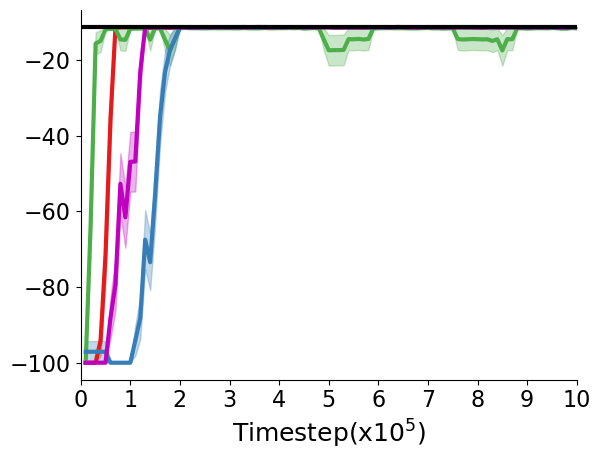}
                \caption{Town02}
        \end{subfigure}
        \caption{\textbf{Learning performance of Deep RL algorithms (Baselines Study)}. Each solid line shows the variation of mean path returns during evaluations, and the bandwidth indicates the standard error of path returns over 30 trials.  Black lines indicate the discounted return of the shortest path in each map.}
    \label{fig:deeprl_performance_returns}
    \vspace{-4mm}
\end{figure}

\begin{figure}
    \centering
    \begin{subfigure}[b]{0.49\columnwidth}
                \centering
                \includegraphics[width=\textwidth]{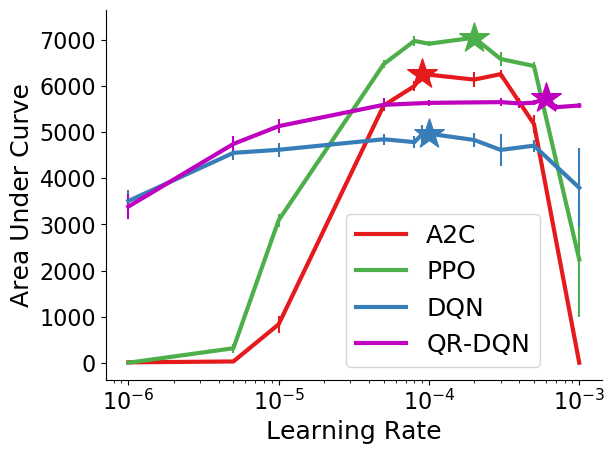}
                \caption{Town01}
        \end{subfigure}
        \begin{subfigure}[b]{0.49\columnwidth}
                \centering
                \includegraphics[width=0.96\textwidth]{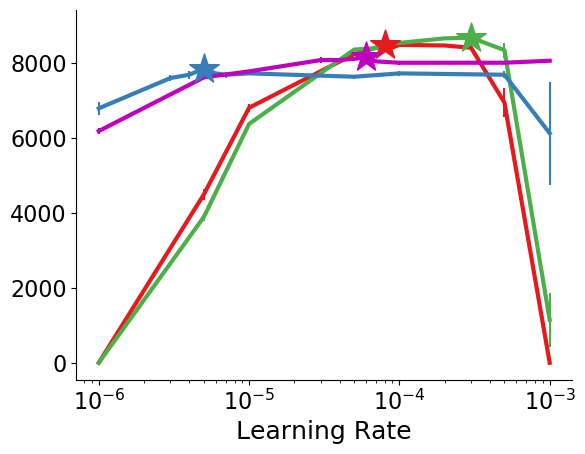}
                \caption{Town02}
        \end{subfigure}
        \caption{\textbf{Sensitivity curves of Deep RL algorithms (Baselines Study)}. Each point in a curve represents the mean value of area under the learning curve over the corresponding five experiments, and the vertical bar at each point reflects the standard error. The points with maximum mean value are marked by stars.}
    \label{fig:deeprl_performance_sensitivity}
    \vspace{-4mm}
\end{figure}

We select four state-of-the-art Deep RL methods, which are A2C \cite{A2C}, PPO \cite{PPO}, DQN \cite{DQN}, and Quantile-Regression DQN (QR-DQN) \cite{dabney2018distributional}, and investigate their performance on the Stochastic Road Network Environment.
In each map, a start state, a pair of goal states (permitting the goal to be reached from either side of a two-way street), and a crosswalk state (designed to lie within the shortest path of the map) are provided.
The base reward $r_{base}=1$, and the penalty on the loopback actions $r_{loopback}=0$.
Agents equipped with the selected methods are trained to learn route planning policies from the start state to a goal state, which are represented by end-to-end multilayer perceptron (MLP) models with the same structure introduced below.
We evaluate a trained agent after every 10,000 steps of learning in the training environment. An evaluation occurs in a separate environment instance, where the agent starts from the same start state and stops when reaching a goal state or 1000 steps.
The curves of discounted return ($\gamma=0.99$) over the trajectories executed during evaluation, shown in Figure \ref{fig:deeprl_performance_returns}, can be used to analyze the learning performance of an agent.

We use implementations of A2C, PPO, DQN and QR-DQN from the Stable-Baselines3 project \cite{stable-baselines3}, and the variable names of hyperparameters used hereafter are the same as in the Stable-Baselines3 code.

\textbf{Hyperparameter selection:} Several hyperparameters were tuned for our experiments, described in detail below.

For A2C, \textit{n\_step} = 64.
This means the agent collects 64 on-policy transitions before updating the policy. 
All collected transitions are used to perform a one-step optimization to update the current policy, and then they are subsequently abandoned. 

For PPO, \textit{n\_step} = 2048, \textit{batch\_size} = 64, and \textit{n\_epochs} = 1. Thus the agent acts for 2048 steps with the current policy, and then the policy is updated by performing a one-step optimization with 64 samples randomly selected from 2048 collected transitions. 
Similar to the A2C agent, all \textit{n\_step} transitions are abandoned after every policy update. 

For DQN, \textit{buffer\_size} = 2048, \textit{batch\_size} = 64, and \textit{gradient\_step} = 1.
In contrast to A2C and PPO, DQN is off-policy, and we use the default $\epsilon$-greedy policy for exploration.
During training, it keeps a buffer of transitions obtained with previous policies and updates the current policy with minibatches from its replay buffer. 
The chosen hyperparameters indicate that a one-step optimization with 64 samples from the buffer containing 2048 previous transitions is performed to update the current policy. In this way, DQN is comparably parameterized with A2C and PPO.

For QR-DQN, we chose \textit{n\_quantiles} = 4, meaning 4 atoms are used to represent the return distribution of a given action.
We also set \textit{exploration\_fraction} = 0.02 and \textit{exploration\_final\_eps} = 0.1, meaning that the exploration rate of the $\epsilon$-greedy policy starts at 1 and decreases linearly to 0.1 during the initial 2\% of total training time steps, then stays at 0.1 thereafter; \textit{buffer\_size} = 2048, \textit{batch\_size} = 64, \textit{gradient\_step} = 1. 


\textbf{Policy network structure:} The route planning policy is represented by a fully-connected multi-layer perceptron (MLP) with four layers.
The input layer takes a vector of the size 65,025 obtained by flattening the observation matrix.
Next, there are two sequential hidden layers, each of which has 64 neurons.
For A2C, PPO and DQN, the output layer gives estimated Q value of actions; for QR-DQN, the output layer gives gives quantile points reflecting the return distribution of actions. 
The action that maximizes estimated Q value or expected value of the return distribution is chosen as the final output.

We study the effect of different learning rates on the performance of the agents in each map.
The range of examined learning rates is from $1\times 10^{-6}$ to $1\times 10^{-3}$.
For each agent and learning rate value, five trials with different random seeds are run, and
the mean area under the learning curves over the trials is computed and used as a metric for evaluating the learning performance at each learning rate -- the results are shown in Figure \ref{fig:deeprl_performance_sensitivity}.
The learning rates corresponding to the maximum mean area under the curve values are marked by stars, and summarized in Table \ref{tab:learning rate}.
Then for each agent in each map, we run another 25 trials with different random seeds, and plot the general performance over 30 trials in Figure \ref{fig:deeprl_performance_returns}.

\begin{table}[h]
    \centering
    \begin{tabular}{c|c|c|c|c}
     & A2C & PPO & DQN & QR-DQN \\ \hline
     Town01 & $9\times 10^{-5}$ & $2\times 10^{-4}$ & $1\times 10^{-4}$ & $6\times 10^{-4}$ \\ \hline
     Town02 & $8\times 10^{-5}$ & $3\times 10^{-4}$ & $5\times 10^{-6}$ & $6\times 10^{-5}$ \\ \hline
    \end{tabular}
    \caption{Best learning rates of Deep RL agents in Baselines Study}
    \label{tab:learning rate}
    \vspace{-2mm}
\end{table}

Four agents learn planning policies that lead to the shortest path within about $5\times 10^5$ steps of experience in most trials in Town01, and about $2\times 10^5$ steps in Town02.
In both maps, A2C and PPO achieve the best learning performance when the learning rate is around $1\times 10^{-4}$, while that of DQN and QR-DQN are not significantly affected by the learning rate. 
As the expected reward of a crosswalk state is $-r_{base}$, which is the same as other states with a deterministic reward, agents learning greedy policies are agnostic to its stochasticity.

\subsection{Robust Route Planning under Uncertainty}
\label{sec:Robust RL}

\begin{figure}
    \centering
        \begin{subfigure}[b]{0.49\columnwidth}
                \centering
                \includegraphics[width=\linewidth]{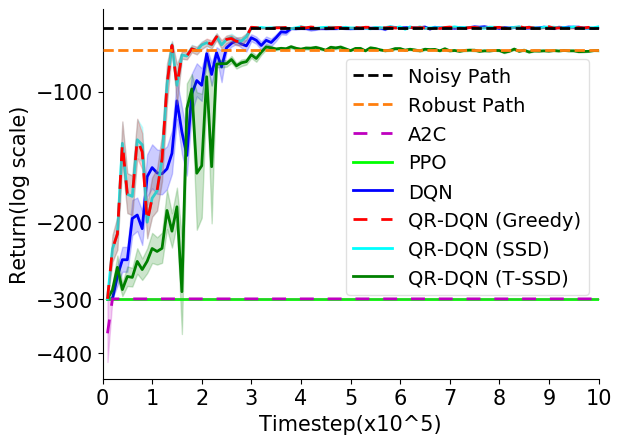}
                \caption{Town01}
        \end{subfigure}
        \begin{subfigure}[b]{0.49\columnwidth}
                \centering
                \includegraphics[width=0.94\linewidth]{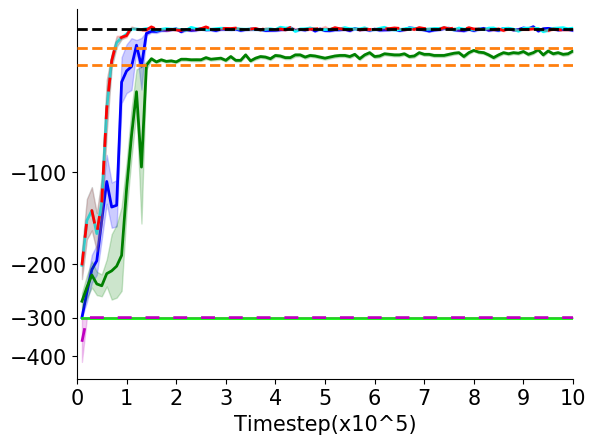}
                \caption{Town02}
                \label{fig:robust_rl_returns_Town02}
        \end{subfigure}
        \caption{\textbf{Learning performance of evaluated algorithms (Robust Route Planning)}. 
        Noisy Path denotes the shortest path, which passes through the crosswalk state. In the Town01 plot, Robust Path denotes the second shortest path, which circumvents the crosswalk state. In the Town02 plot, an additional Robust Path is also shown, which is actually the third shortest path. 
        }
    \label{fig:robust_rl_returns}
    \vspace{-4mm}
\end{figure}

\begin{figure*}[th!]
    \centering
        \includegraphics[width=0.75\textwidth]{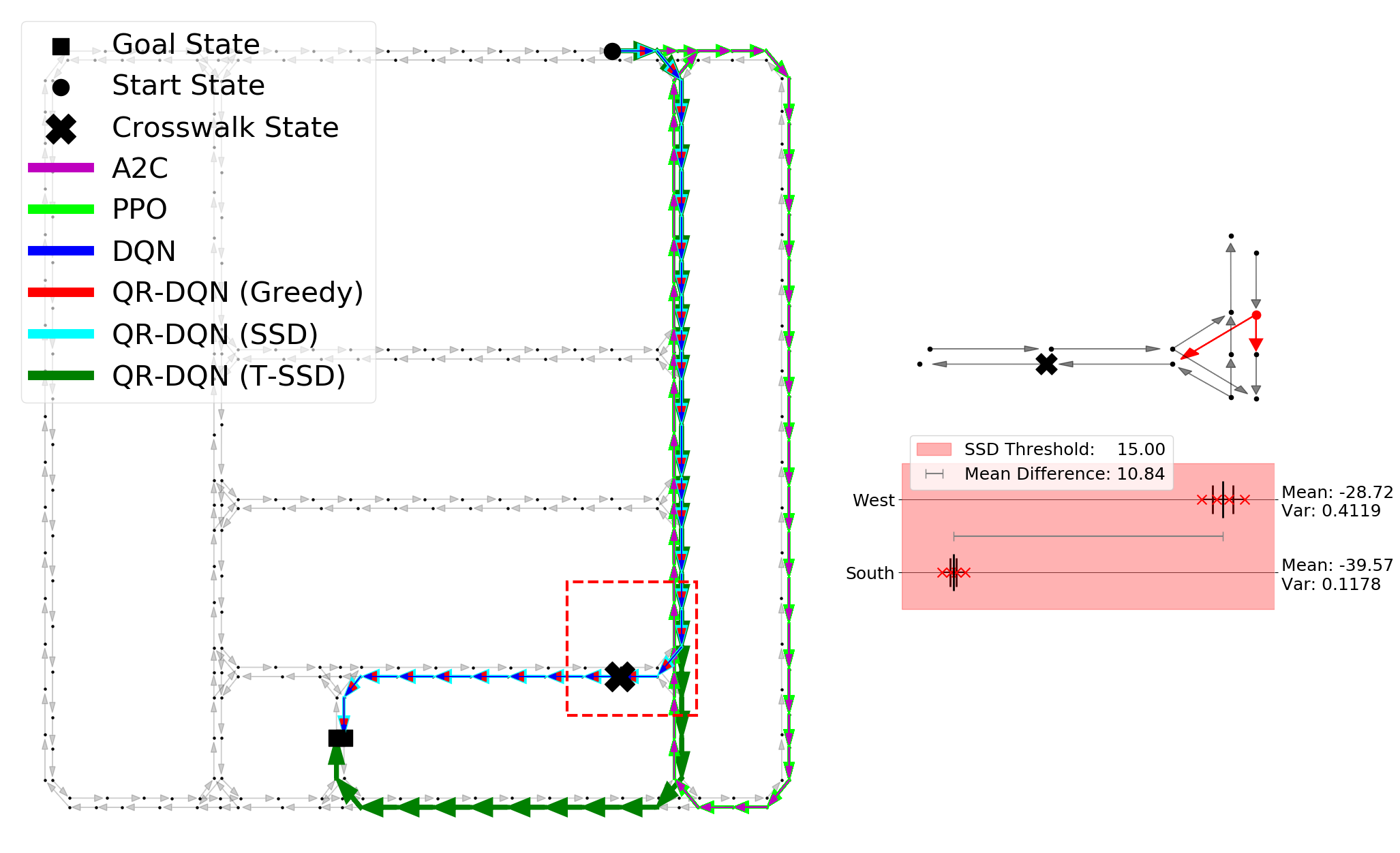}
        \caption{\textbf{Planned routes of evaluated algorithms during the final evaluation in Town01 (Robust Route Planning)}. We also zoom in on the area where the Noisy path and the Robust Path diverge (Red rectangle box), and draw the critical state and its actions in red. In the action return distributions, red crosses mark learned quantiles, the central vertical bar marks the mean, and the distance between the central bar and a lateral short bar equals the variance.
        }
    \label{fig:Town01_robust_rl_paths}
    \vspace{-4mm}
\end{figure*}

\begin{figure*}[th!]
    \centering
        \includegraphics[width=0.85\textwidth]{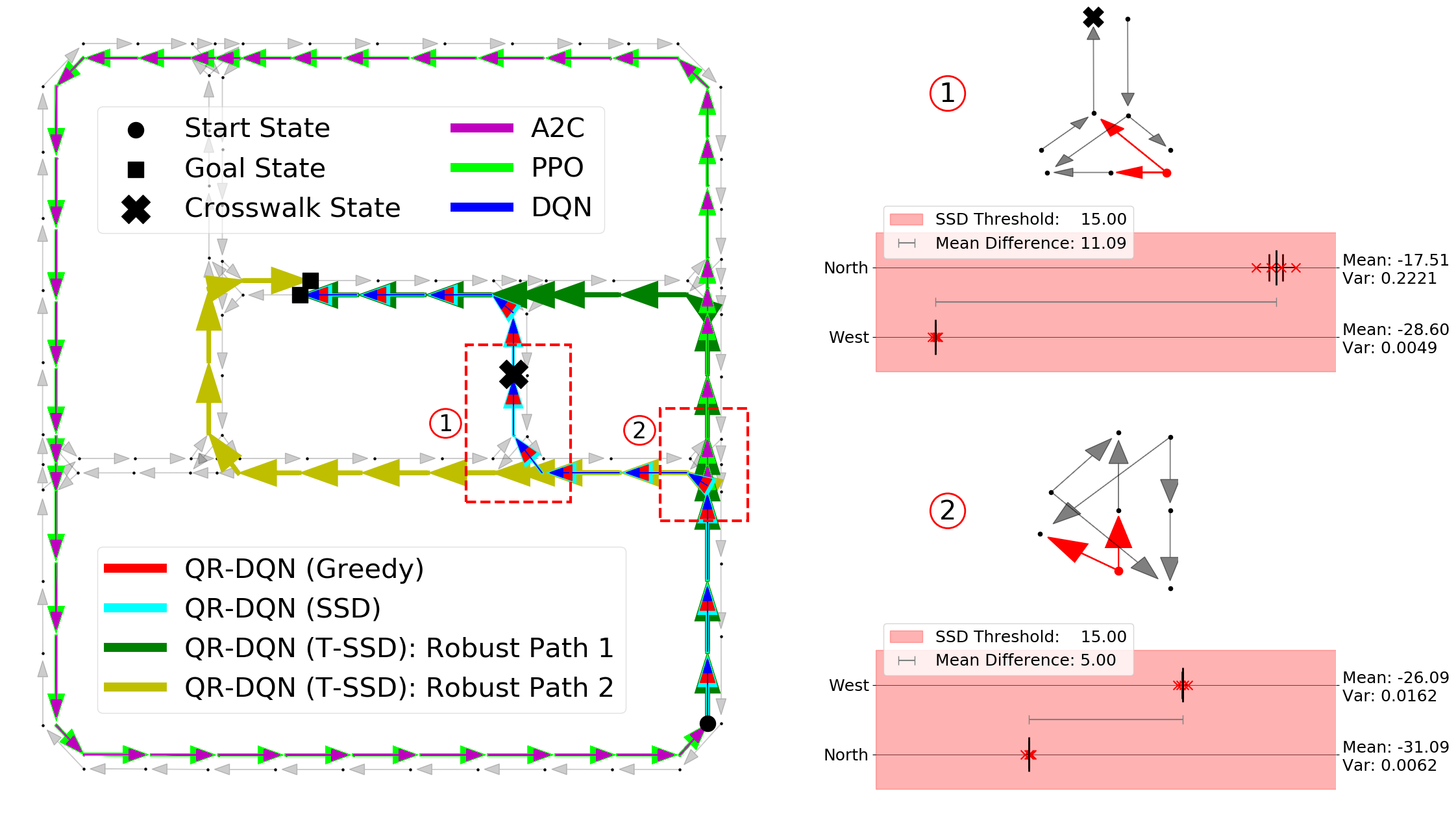}
     \hfill
        \caption{\textbf{Planned routes of evaluated algorithms during the final evaluation in Town02 (Robust Route Planning)}. QR-DQN (T-SSD) plans the second shortest path (Robust Path 1) in 25 trials, and it selects the third shortest path (Robust Path 2) in 5 trials.
        The action return distributions of critical states 1 and 2 come from a trial that chooses Robust Path 1.
        }
    \label{fig:Town02_robust_rl_paths}
    \vspace{-4mm}
\end{figure*}



Based on the results of Section \ref{sec:baselines}, the learning rate we choose for each agent is shown in Table \ref{tab:robust_rl}, and other hyperparameters remain the same as those in Section \ref{sec:baselines}.
We also include the proposed Robust Learning and Execution framework into the experiment, with three execution policies: (1) Greedy (default policy in QR-DQN); (2) SSD; (3) Thresholded SSD (denoted as T-SSD in figures). For the Thresholded SSD policy, we choose $ssd\_thres = 5\cdot r_{base}$.

\begin{table}[h]
    \centering
    \begin{tabular}{c|c|c|c}
        A2C & PPO & DQN & QR-DQN  \\ \hline
        $1\times 10^{-4}$ & $1\times 10^{-4}$ & $1\times 10^{-4}$ & $5\times 10^{-4}$ \\ \hline
    \end{tabular}
    \caption{Learning rates of Deep RL agents in Robust Route Planning Study}
    \label{tab:robust_rl}
    \vspace{-2mm}
\end{table}

We use the same start state, goal states and crosswalk state as those used in the baselines study.
The base reward $r_{base}=3$.
As we are concerned with the effect of crosswalk stochasticity on route planning, a positive penalty on the loopback actions $r_{loopback}=18$ is used.
In this experiment, we use a Convolutional Neural Network (CNN) model for the planning policies.
Similar to the baselines study, evaluation is performed in a separate environment after every 10,000 steps of learning.
We run 30 trials with different seeds for each map, and plot the overall performance of each algorithm in Figure \ref{fig:robust_rl_returns}.
We also plot the planned route of each algorithm during the final evaluation in Figure \ref{fig:Town01_robust_rl_paths} and Figure \ref{fig:Town02_robust_rl_paths}. 


\textbf{Policy network structure:} The route planning policy is represented by a Convolutional Neural Network (CNN) model with the following structure.
The input is the observation matrix, of size $255\times 255 \times 1$.
The first hidden layer convolves 16 filters of the size $8\times 8$ with stride 4 with the input, followed by a rectifier.
The second hidden layer convolves 32 filters of the size $4\times 4$ with stride 2, followed by a rectifier.
The third hidden layer convolves 64 filters of the size $4\times 4$ with stride 2, followed by a rectifier.
The fourth hidden layer convolves 64 filters of the size $3\times 3$ with stride 1, followed by a rectifier.
Two sequential fully-connected layers, each of which has 64 neurons, follow that.
Similarly, the output layer gives the estimated Q values of actions (A2C, PPO and DQN) or quantile points reflecting the return distribution of actions (QR-DQN).
Note that for a QR-DQN agent, the output action is determined by its execution policy.

It can be seen that the results over the two maps are similar.
We note that when equipped with the Thresholded SSD policy, our proposed framework is the only one that learns to execute a high-quality robust route, which avoids travel time stochasticity and is still short in length.
In Town02, the second and the third shortest paths are both free of stochasticity, and very close in length, thus it is relatively slow for the algorithm to learn to move from the third shortest path to the second second path.
We denote the second and the third shortest path in Town02 as Robust Path 1 and Robust Path 2, respectively. 

For our Town01 results (shown in Figure \ref{fig:Town01_robust_rl_paths}), besides the planned paths, we also show the return distributions of actions (learned by QR-DQN) at the critical state after which the Noisy path and the Robust Path diverge.
It can be seen that the action leading to the Noisy path has a higher expected return but a greater variance, hence the Thresholded SSD policy favors the other action with less variance, while the expected return is still within the SSD threshold.

For our Town02 results (shown in Figure \ref{fig:Town02_robust_rl_paths}), we draw the action return distributions for two critical states.
Critical state \textcircled{1} decides whether to take the Robust Path 2 or the Noisy Path, and critical state \textcircled{2} decides whether to take Robust Path 1 or the other two.
It could be seen that at critical state \textcircled{1}, the difference in variances is large, hence the proposed framework quickly learns to avoid taking the Noisy path.
However, critical state \textcircled{2} is more distant from and less affected by the crosswalk state, and the variance of the final learned distributions of both actions are quite close and small.
Hence, this explains why the learning curve of QR-DQN (T-SSD) in Figure \ref{fig:robust_rl_returns_Town02} converges slowly from Robust Path 2 to Robust Path 1.

The Greedy policy is agnostic to environment stochasticity, so it takes the Noisy Path in all trials.
Because there are no exact ties in expectation, the SSD policy exhibits identical behavior to Greedy.
As the Thresholded SSD policy plans a robust path at the price of a longer expected travel time, the user may select the stricter SSD policy if robustness is less preferred. 
DQN also learns a greedy policy that selects the shortest paths in both maps. 

However, A2C and PPO fail to learn a route to the goal in both maps, and it can be seen that they become stuck in loops. It is likely because they are on-policy algorithms, which may converge to a local optimal solution before sufficient exploration. Additionally, the penalty on the loopback action also leads to bias towards the other action and contributes to A2C and PPO converging to locally optimal solutions.

\section{Conclusion}
\label{sec:conclusion}

This paper introduces the Robust Learning and Execution framework for the route planning problem of mobile robots in unknonwn and stochastic environments.  
The framework learns environmental stochasticity with QR-DQN, and could use different execution policies according to user preferences on route robustness.
We use the SSD policy, which is based on the second-order stochastic dominance (SSD) relation, to determine robust actions during execution, and we introduce a relaxation, the Thresholded SSD policy, that allows further preferences on robustness.
Experimental results on a simulated road network environment show that our proposed framework could achieve adjustable performance with different execution policies: (1) Due to no strict tie in expected returns, the SSD policy selects routes that minimize expected travel time but with higher stochasticity; (2) When equipped with the Thresholded SSD policy, our framework could select short robust routes with minimum stochasticity in travel time, at the cost of higher expected travel time.

\section*{Acknowledgments}
This research was supported by the Office of Naval Research, grant N00014-20-1-2570.

\bibliographystyle{elsarticle-num}
\bibliography{main}

\end{document}